\documentclass{article}

\usepackage{PRIMEarxiv}

\usepackage[utf8]{inputenc} 
\usepackage[T1]{fontenc}    
\usepackage{hyperref}       
\usepackage{url}            
\usepackage{booktabs}       
\usepackage{amsfonts}       
\usepackage{amsmath}        
\usepackage{nicefrac}       
\usepackage{microtype}      
\usepackage{fancyhdr}       
\usepackage{graphicx}       
\usepackage{float}
\usepackage{caption}
\usepackage{algorithmic}
\usepackage{algorithm}

\title{4D SlingBAG: spatial-temporal coupled Gaussian ball for large-scale dynamic 3D photoacoustic iterative reconstruction}

\author{
  Shuang Li* \\
  College of Future Technology \\
  Peking University \\
  Beijing, China \\
  \texttt{jaeger\_ls@stu.pku.edu.cn} \\
  \And
  Yibing Wang* \\
  College of Future Technology \\
  Peking University \\
  Beijing, China \\
  \texttt{ddffwyb@pku.edu.cn} \\
  \And
  Jian Gao* \\
  School of Intelligence Science and Technology \\
  Nanjing University \\
  Suzhou, China \\
  \texttt{jian\_gao@smail.nju.edu.cn} \\
  \And
  Chulhong Kim \\
  Department of Electrical Engineering, Convergence IT Engineering, Mechanical Engineering, \\
  and Medical Science and Engineering, Medical Device Innovation Center, \\
  Pohang University of Science and Technology (POSTECH), \\
  77 Cheongam-ro, Nam-gu, Pohang, Gyeongbuk 37673, Republic of Korea \\
  \texttt{chulhong@postech.edu} \\
  \And
  Seongwook Choi \\
  Department of Electrical Engineering, Convergence IT Engineering, Mechanical Engineering, \\
  and Medical Science and Engineering, Medical Device Innovation Center, \\
  Pohang University of Science and Technology (POSTECH), \\
  77 Cheongam-ro, Nam-gu, Pohang, Gyeongbuk 37673, Republic of Korea \\
  \texttt{swchoi715@postech.ac.kr} \\
  \And
  Yu Zhang \\
  College of Future Technology \\
  Peking University \\
  Beijing, China \\
  \texttt{zyuaiyi1\_@stu.pku.edu.cn} \\
  \And
  Qian Chen \\
  College of Future Technology \\
  Peking University \\
  Beijing, China \\
  \texttt{chen\_qian@stu.pku.edu.cn} \\
  \And
  Yao Yao** \\
  School of Intelligence Science and Technology \\
  Nanjing University \\
  Suzhou, China \\
  \texttt{yaoyao@nju.edu.cn} \\
  \And
  Changhui Li** \\
  College of Future Technology \\
  Peking University \\
  Beijing, China \\
  \texttt{chli@pku.edu.cn} \\
}

\begin{document}

\maketitle

\newcommand\nnfootnote[1]{%
  \begin{NoHyper}
  \renewcommand\thefootnote{}\footnote{#1}%
  \addtocounter{footnote}{-1}%
  \end{NoHyper}
}
\nnfootnote{* Co-first author. ** Corresponding author.}

\begin{abstract}
Large-scale dynamic three-dimensional (3D) photoacoustic imaging (PAI) is significantly important in clinical applications. In practical implementations, large-scale 3D real-time PAI systems typically utilize sparse two-dimensional (2D) sensor arrays with certain angular deficiencies, necessitating advanced iterative reconstruction (IR) algorithms to achieve quantitative PAI and reduce reconstruction artifacts. However, for existing IR algorithms, multi-frame 3D reconstruction leads to extremely high memory consumption and prolonged computation times, with limited consideration of the spatial-temporal continuity between data frames. Here, we propose a novel method, named the 4D sliding Gaussian ball adaptive growth (4D SlingBAG) algorithm, based on the current point cloud-based IR algorithm sliding Gaussian ball adaptive growth (SlingBAG), which has minimal memory consumption among IR methods. Our 4D SlingBAG method applies spatial-temporal coupled deformation functions to each Gaussian sphere in point cloud, thus explicitly learning the deformations features of the dynamic 3D PA scene. This allows for the efficient representation of various physiological processes (such as pulsation) or external pressures (e.g., blood perfusion experiments) contributing to changes in vessel morphology and blood flow during dynamic 3D PAI, enabling highly efficient IR for dynamic 3D PAI. Simulation experiments demonstrate that 4D SlingBAG achieves high-quality dynamic 3D PA reconstruction. Compared to performing reconstructions by using SlingBAG algorithm individually for each frame, our method significantly reduces computational time and keeps a extremely low memory consumption. The project for 4D SlingBAG can be found in the following GitHub repository: \href{https://github.com/JaegerCQ/4D-SlingBAG}{https://github.com/JaegerCQ/4D-SlingBAG}.

\end{abstract}


\section{Introduction}

Photoacoustic imaging (PAI) uniquely integrates ultrasound detection and optical absorption contrast, making it a powerful label-free optical imaging modality capable of imaging living tissue several centimeters deep with high spatial resolution~\cite{park2024clinical, wang2012photoacoustic, assi2023review, dean2017advanced, lin2022emerging, ntziachristos2024addressing}. PAI enables 3D imaging of biological tissues, and recent advancements in spherical and planar arrays for three-dimensional imaging~\cite{matsumoto2018label, matsumoto2018visualising, ivankovic2019real, dean2013portable, nagae2018real, kim2023wide, piras2009photoacoustic, heijblom2012visualizing} have spurred the development of 3D PAI, especially for clinical applications. However, in practical disease diagnosis and functional studies, dynamic tissue changes and hemodynamics are often of greater interest~\cite{ahn2021high, mantri2022monitoring}. This necessitates the development of high-quality dynamic PAI techniques with temporal resolution.

Fortunately, most 3D PAI systems are capable of acquiring dynamic 3D data. However, due to the high cost and complexity of system construction, these devices often employ sparsely arranged 2D arrays. This requires utilizing advanced iterative reconstruction (IR) algorithms instead of universal back-projection (UBP) methods to reconstruct the images and reduce reconstruction artifacts caused by a limited number of transducer elements. Currently, most work in dynamic 3D PAI requires frame-by-frame reconstruction~\cite{xiang20134}, neglecting the temporal continuity of tissue and blood flow states across frames. Only several works on IR algorithms view the dynamic PA scene as whole~\cite{lozenski2024proxnf}, while those methods are based on voxel grid neural field representation, results in excessive memory consumption and reconstruction time. This shortcoming severely limits both reconstruction efficiency and image quality, especially for large-scale 3D reconstructions where the memory consumption for processing multiple frames exceeds the capabilities of most GPUs.

SlingBAG~\cite{li2024sliding} introduces a point-cloud-based IR framework, explicitly representing the PA scene using a series of Gaussian spheres with varying standard deviations, initial acoustic pressures, and 3D coordinates. Compared to conventional iterative methods, SlingBAG significantly reduces memory consumption by orders of magnitude, enabling efficient large-scale 3D PAI reconstruction. However, this method is currently designed solely for single-frame 3D reconstruction. For multi-frame PA datasets, each frame must be reconstructed independently. To improve the efficiency of multi-frame reconstructions, it is necessary to introduce temporal deformation functions that couple the deformation of each Gaussian sphere in SlingBAG both spatially and temporally to achieve efficient dynamic 3D PA iterative reconstruction.

The task of dynamic 3D scene reconstruction is also a vital issue in computer vision. One of the most efficient 3D scene reconstruction algorithms in computer vision, 3D Gaussian Splatting~\cite{kerbl20233Dgaussians}, similarly faces the limitation of being tailored only for static single-frame scenarios, yet is not readily applicable for dynamic multi-frame reconstructions. To extend 3D Gaussian Splatting for efficient dynamic 3D scene tasks, researchers in computer vision have explored several solutions, including 4D Gaussian Splatting, which uses 4D neural voxel encoding and multiple multilayer perceptrons (MLPs) for decoding~\cite{wu20244d}, as well as Gaussian-Flow~\cite{lin2024gaussian} and Deform3DGS~\cite{yang2024deform3dgs}, which leverage explicit spatial-temporal deformation functions to model the deformation of 3D Gaussian ellipsoids. Notably, Deform3DGS employs learnable Gaussian functions as the basis for deformation function, achieving high representational capacity and efficiency with a small number of parameters.

Inspired by Deform3DGS, we propose the 4D SlingBAG algorithm, which introduces an explicit deformation model for each Gaussian sphere in the point cloud representation of SlingBAG. Designed specifically for PAI, this approach avoids the need for complex neural network structures, enabling efficient reconstruction of dynamic 3D PA scenes while maintaining the low memory consumption and fast iterative reconstruction speed of SlingBAG. Specifically, we represent the dynamic 3D PA source as a set of deformable Gaussian spheres and employ learnable Gaussian functions as the basis for the spatial-temporal deformation function. This allows explicit modeling of the deformation of each Gaussian sphere's attributes, including amplitude (initial acoustic pressure), standard deviation (size), and mean value (spatial coordinates), forming a Gaussian-deformed Gaussian ball model (GGBall model). The GGBall model provides a compact representation of the dynamic PA scene, requiring only a single point cloud and the corresponding deformation parameters to achieve full dynamic 3D PAI reconstruction. Results demonstrate that, compared to frame-by-frame SlingBAG reconstructions, the proposed 4D SlingBAG based on the GGBall model achieves more than $\times 8$ iterative reconstruction speed for dynamic 3D PAI, and the memory consumption for dynamic 3D reconstruction using 4D SlingBAG is almost the same as that for single-frame 3D reconstruction with SlingBAG.

\section{Method}

In this section, we present the 4D SlingBAG algorithm for dynamic 3D PAI reconstruction. First, in Section 2.1, we review the SlingBAG algorithm used for large-scale 3D PAI reconstruction. Then, in Section 2.2, we provide a detailed description of using spatial-temporal coupled deformation functions to represent dynamic PA scenes, explicitly modeling the motion of reconstruction targets through the Gaussian-deformed Gaussian ball model (GGBall).

\subsection{Recap on SlingBAG algorithm}

The SlingBAG algorithm is the first method that stores information of PA source in the form of point cloud that undergoes iterative optimization, wherein the 3D PA scene is modeled as a series of Gaussian-distributed spherical sources. During the iterative reconstruction process, all these Gaussian-distributed sources with specific peak intensity $p_0$ (pressure value), standard deviation $a_0$ (size), and mean value $\mu$ (spatial position $\left(x_s, y_s, z_s\right)$) are used to calculate the predicted PA signals in the position of sensors. By minimizing the discrepancies between the predicted and the actual PA signals, the SlingBAG algorithm iteratively refine the point cloud and ultimately realize the 3D PAI reconstruction by converting the reconstruction result from point cloud into voxel grid.

\subsection{Spatial-temporal
coupled deformation: 4D SlingBAG with Gaussian-deformed Gaussian ball model}

Dynamic 3D PAI often involves tissue deformation and blood flow changes induced by pulsation. During this process, the 3D morphology and optical absorption properties of the reconstructed tissue undergo changes. In the point cloud model, this implies that the peak intensity, the standard deviation and 3D coordinates of each Gaussian sphere, representing PA sources, will vary over time. Given that consecutive temporal frames exhibit both certain differences and similarities, employing a spatial-temporal coupled deformation function allows simultaneous consideration of both the static and dynamic components of the scene. This enables efficient reconstruction of dynamic 3D PAI.

\begin{figure}[ht]
    \centering
    \includegraphics[width=0.8\linewidth]{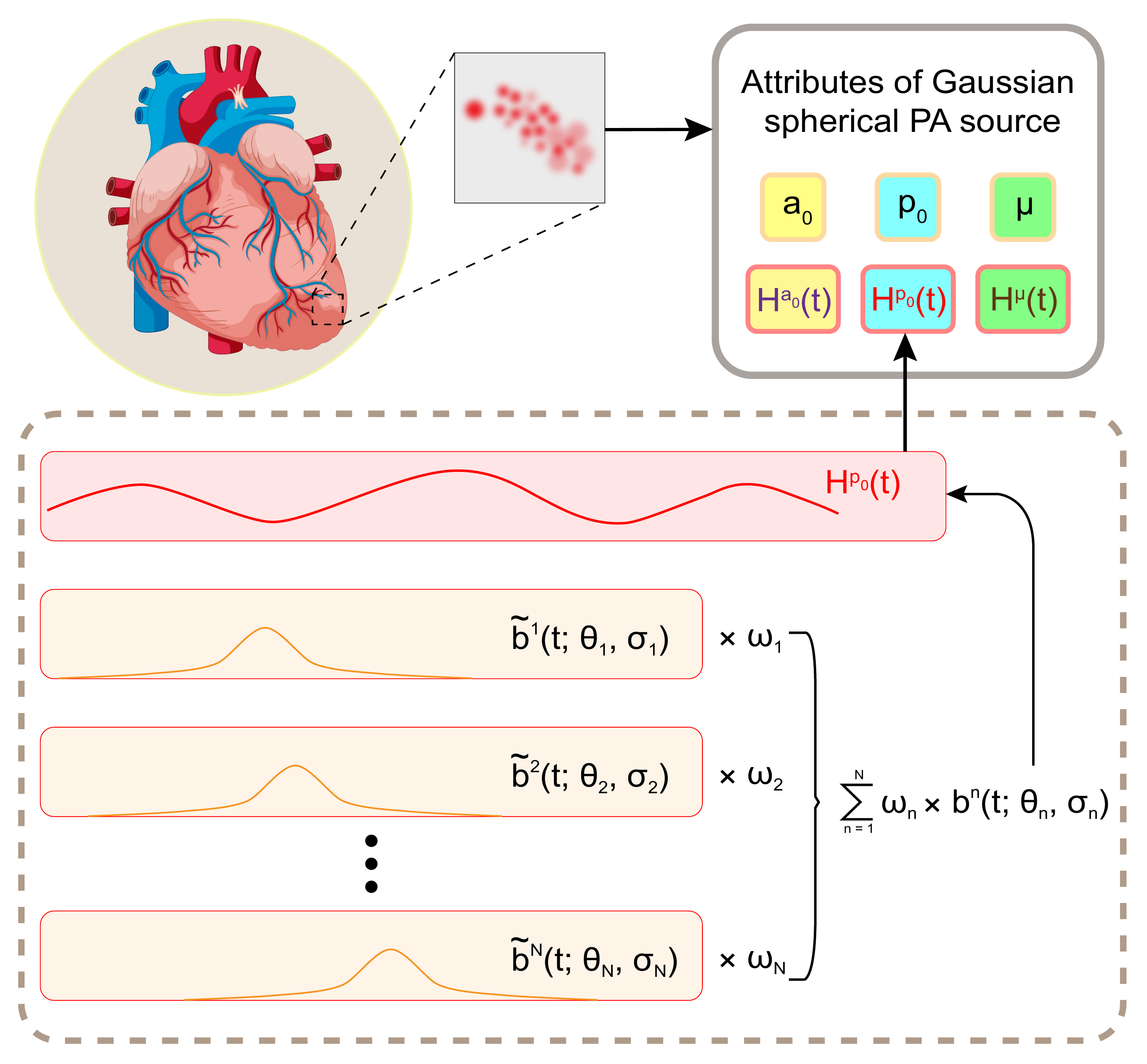}
    \caption{The Gaussian-deformed Gaussian ball model (GGBall).}
    \label{fig:fig1}
\end{figure}

The aim of 4D SlingBAG is to directly learn the dynamic characteristics of each Gaussian sphere PA source with a spatial-temporal coupled deformation function. Inspired by Deform3DGS, we similarly utilize Gaussian basis functions to construct the spatial-temporal coupled deformation function $H(t)$. This ensures continuity in deformations across adjacent temporal frames while maintaining near-independence for frames with large temporal separations. Such a design provides high computational efficiency and the capacity to capture complex deformation dynamics~\cite{yang2024deform3dgs}. However, we tailored this approach specifically for dynamic PA scenarios, allowing each Gaussian sphere's standard deviation, 3D coordinates, and initial acoustic pressure to independently learn their respective deformation dynamics. We refer to this approach as the Gaussian-deformed Gaussian ball model (GGBall), as demonstrated in Fig.~\ref{fig:fig1}.

In the GGBall model, the attributes of each Gaussian sphere PA source at different time frames consist of two components, as shown in Eq.~\ref{eq:eq1}: one is the static baseline attribute $\left(a_0, p_0, \mu\right)$ initialized from a fixed reference frame, and the other is a transient residual value associated with the spatial-temporal coupled deformation function $H(t)$. 

\begin{equation}
\begin{split}
    a_0(t)&=a_0(1+H^{a_0}_N(t)),\\
    \quad \\
    p_0(t)&=p_0(1+H^{p_0}_N(t)),\\
    \quad \\
    \mu(t)&=\mu(1+H^{\mu}_N(t)).
    \label{eq:eq1}
\end{split}
\end{equation}

The transient residual takes as input the normalized time frame and outputs the deformation value of each attribute at different temporal frames. The deformation function $H(t)$ uses Gaussian functions with varying centers $\theta$ and variances $\sigma$, combined with different weights $\omega$ as basis, as shown in Eq.~\ref{eq:eq2}. Both the baseline static attributes of each Gaussian sphere $\left(a_0, p_0, \mu\right)$ and the parameters constructing the spatial-temporal coupled deformation function — $\left(\theta, \sigma, \omega\right)$ of each basis — are fully learnable.

\begin{equation}
\begin{split}
    H_N(t) = \sum_{n=1}^{N}\omega_n\tilde{b}^n\left(t; \theta_n, \sigma_n\right),\\
    \text{where}\quad\tilde{b}^n\left(t; \theta_n, \sigma_n\right) = \exp\left(-\frac{1}{2\sigma_n^2}\left(t-\theta_n\right)^2\right).
    \label{eq:eq2}
\end{split}
\end{equation}

\begin{figure}[ht]
    \centering
    \includegraphics[width=0.9\linewidth]{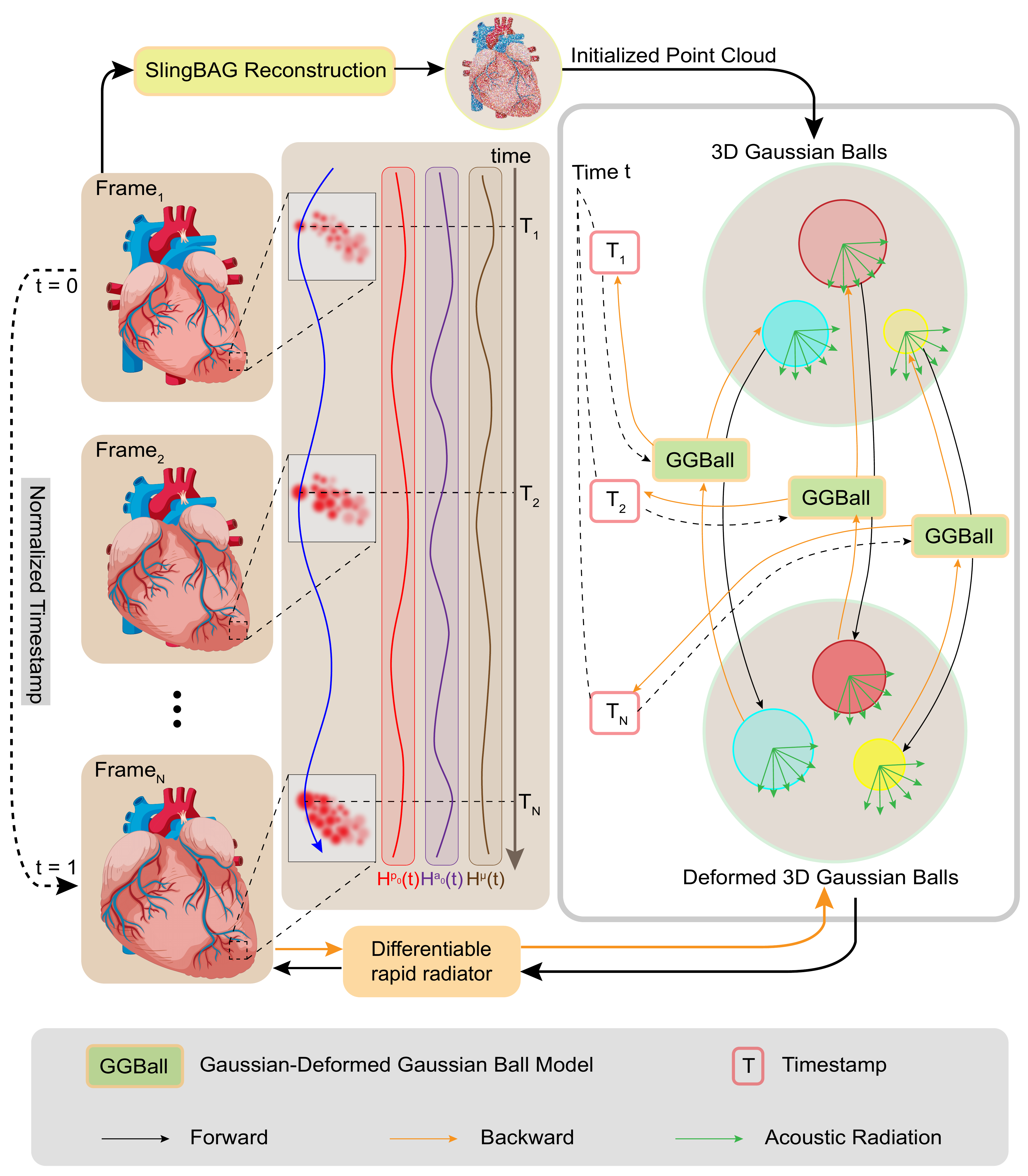}
    \caption{The overview framework of 4D SlingBAG pipeline.}
    \label{fig:fig2}
\end{figure}

The dynamic 3D PA iterative reconstruction process of 4D SlingBAG can be divided into two steps, as shown in Fig.~\ref{fig:fig2}. First, a specific time frame is randomly selected as the reference time frame, and SlingBAG is used to perform a initial reconstruction for this time frame. The reconstruction result from SlingBAG is then used as the initialized baseline static attributes of the point cloud input for 4D SlingBAG, and the spatial-temporal coupled deformation function is initialized simultaneously.

Next, using SlingBAG's differentiable rapid radiator~\cite{li2024sliding}, the forward process is computed for each Gaussian sphere PA source based on its time-varying attributes $\left(a_0(t), p_0(t), \mu(t)\right)$. The computed forward results are compared with the real data of the corresponding time frames to calculate the $L_2$ loss. This process iteratively updates all baseline static attributes of the Gaussian spheres as well as the parameters of their corresponding spatial-temporal coupled deformation functions $H(t)$. Through this iterative process, the dynamic 3D PA scene is reconstructed effectively.

\section{Results}

\subsection{Simulation experiment 1: simple phantom reconstruction}

To validate the dynamic 3D PAI reconstruction capability of 4D SlingBAG, we conducted a simulation based on a pulsating heart model. The detector array was configured as a spherical array with a radius of 60 mm, containing 1,024 detector elements arranged across the entire sphere surface following a Fibonacci sequence layout. The PA source was a "simulated heart" located within a cubic region of $x, y, z \in \left(-25.6\text{ mm}, 25.6\text{ mm}\right)$ inside the spherical array.

\begin{figure}[ht]
    \centering
    \includegraphics[width=0.9\linewidth]{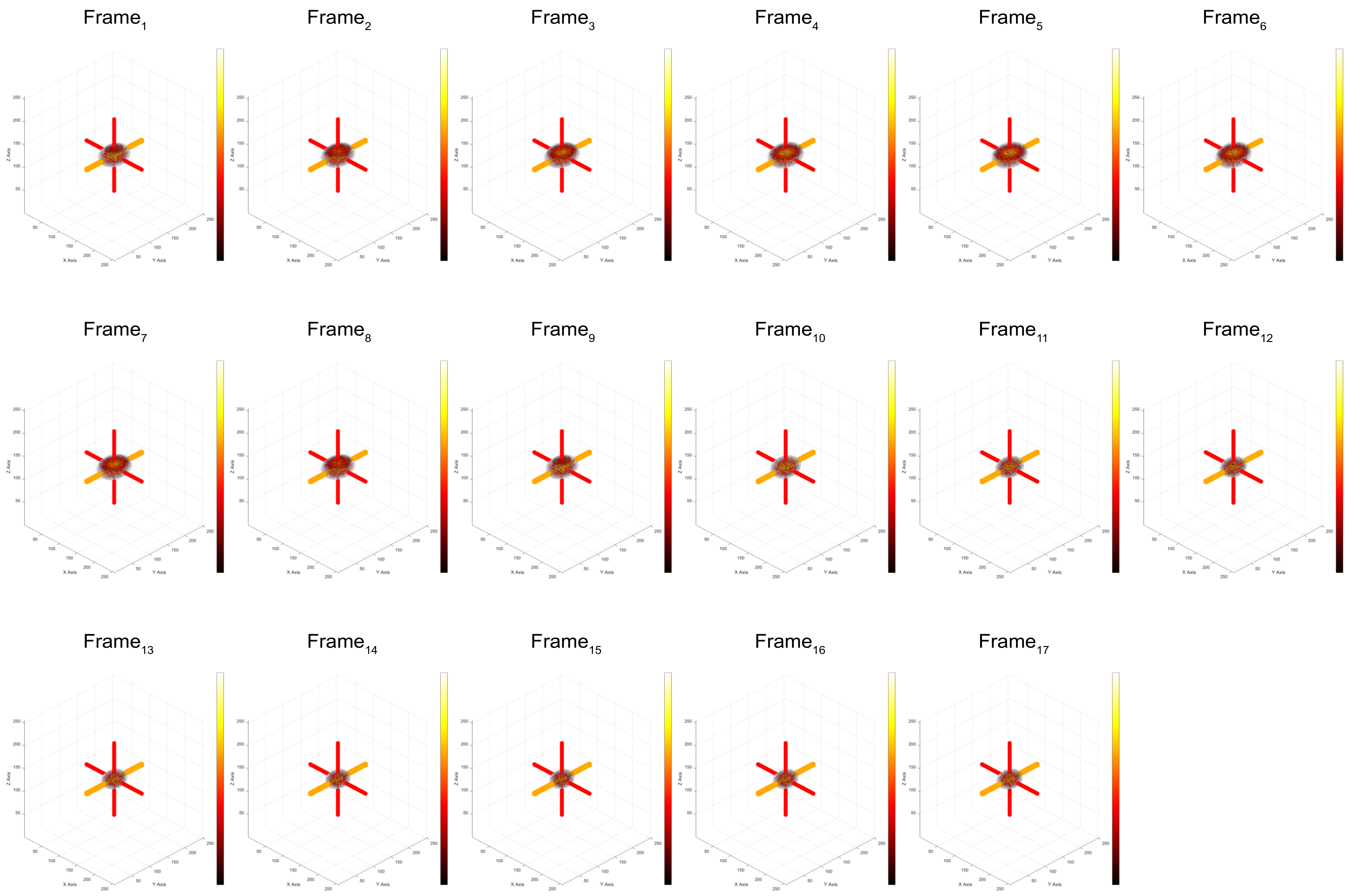}
    \caption{The dynamic 3D photoacoustic scene.}
    \label{fig:fig3}
\end{figure}

The simulated heart consisted of three mutually perpendicular rectangular regions with uniform acoustic pressure distributions and a Gaussian ellipsoid region located at the center. While the rectangular regions remained stationary, the acoustic pressure and size of the Gaussian ellipsoid dynamically changed over time to mimic the pulsation of a heart. The entire simulated dynamic 3D PA scene was constructed across 17 temporal frames (Fig.~\ref{fig:fig3}).

\begin{figure}[H]
    \centering
    \includegraphics[width=0.9\linewidth]{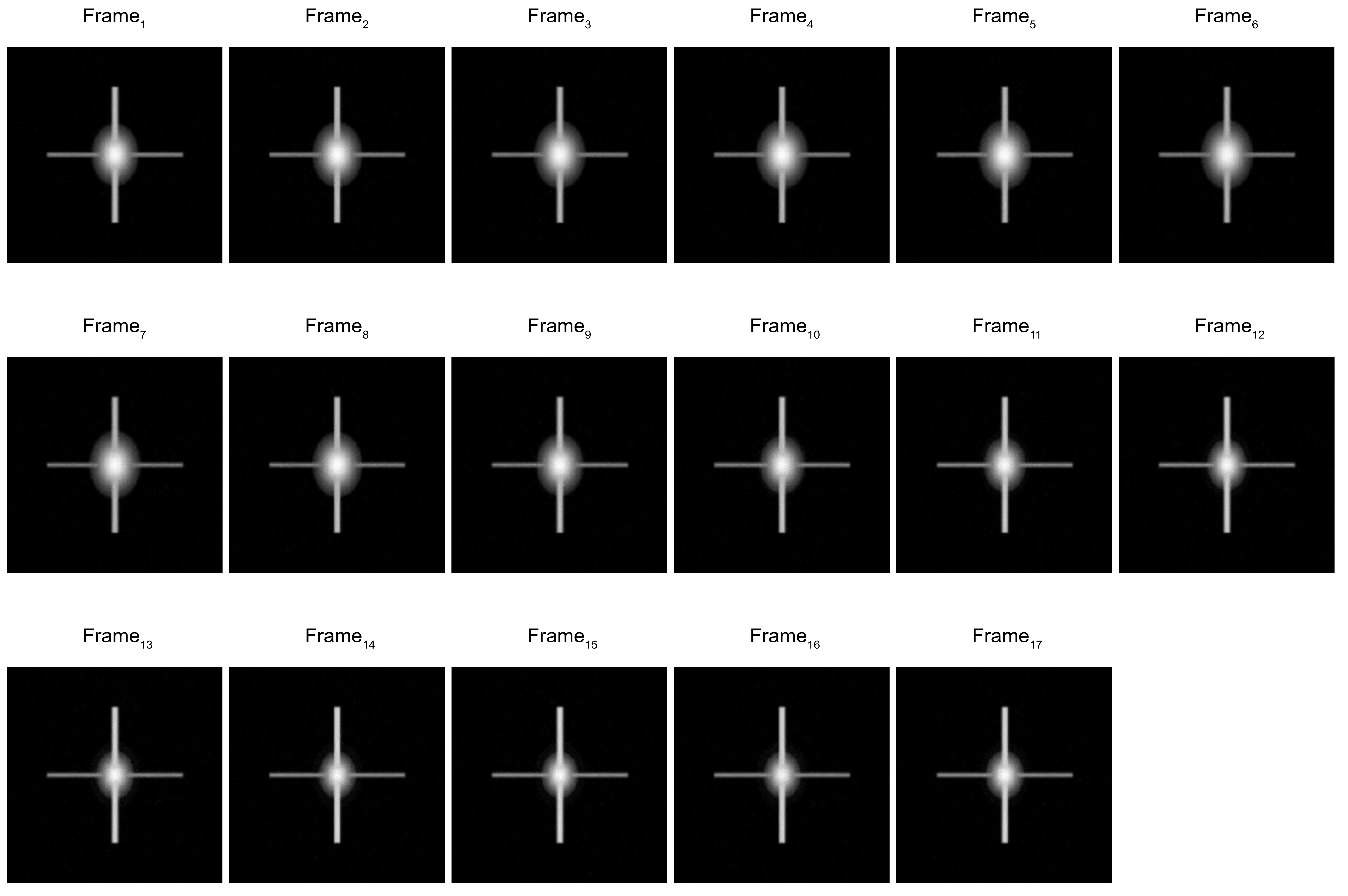}
    \caption{The Top-View-MAP of the dynamic 3D photoacoustic reconstruction results using 4D SlingBAG.}
    \label{fig:fig4}
\end{figure}

\begin{figure}[H]
    \centering
    \includegraphics[width=0.9\linewidth]{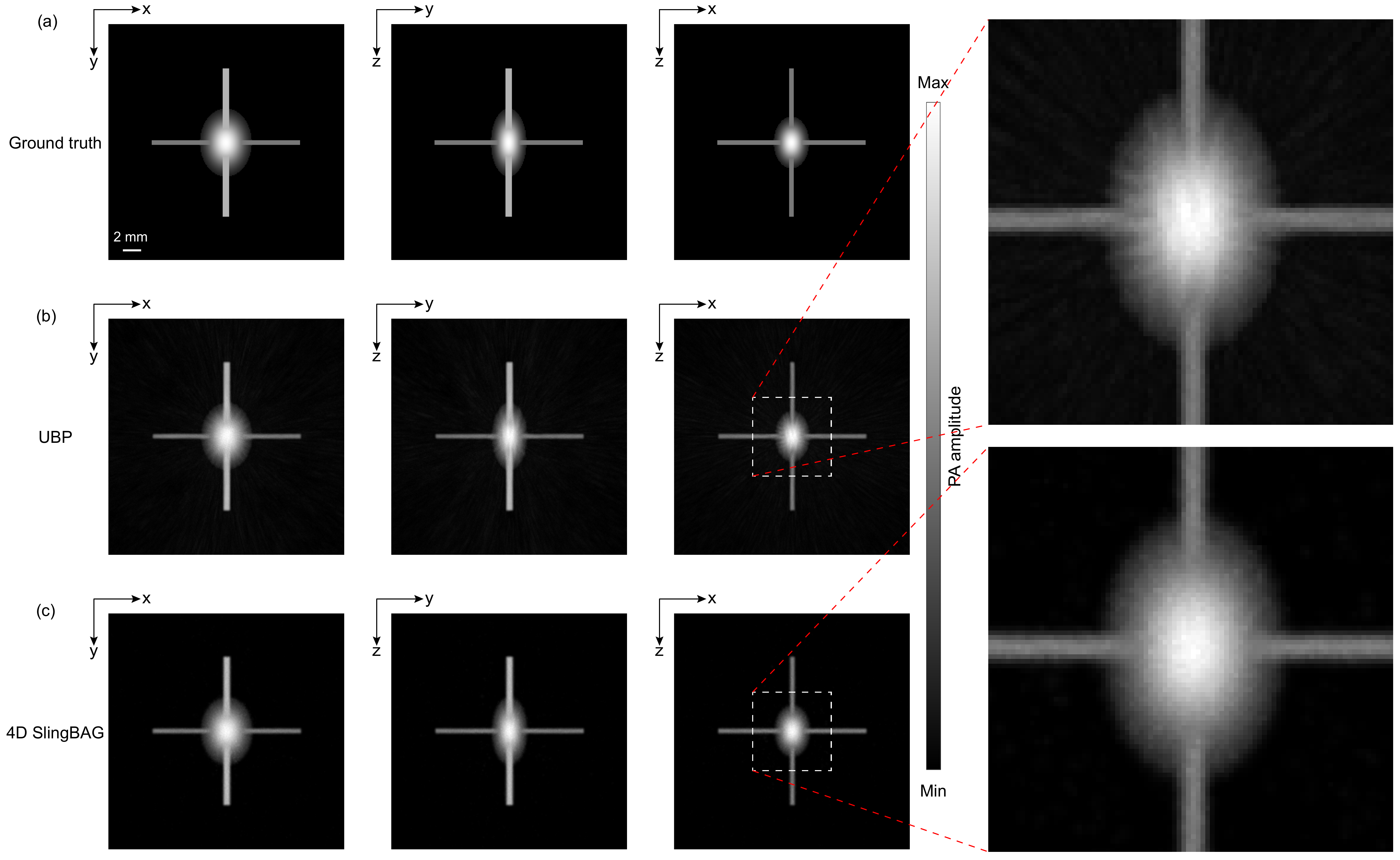}
    \caption{The reconstruction results of the dynamic 3D photoacoustic scene in the 1st frame. (a) XY Plane-MAP, YZ Plane-MAP, XZ Plane-MAP of the ground truth acoustic source. (b) XY Plane-MAP, YZ Plane-MAP, XZ Plane-MAP of the UBP reconstruction results using 1,024 sensor signals. (c) XY Plane-MAP, YZ Plane-MAP, XZ Plane-MAP of the 4D SlingBAG reconstruction results using 1,024 sensor signals. (Scale: 2.0 mm.)}
    \label{fig:fig5}
\end{figure}

\begin{figure}[H]
    \centering
    \includegraphics[width=0.9\linewidth]{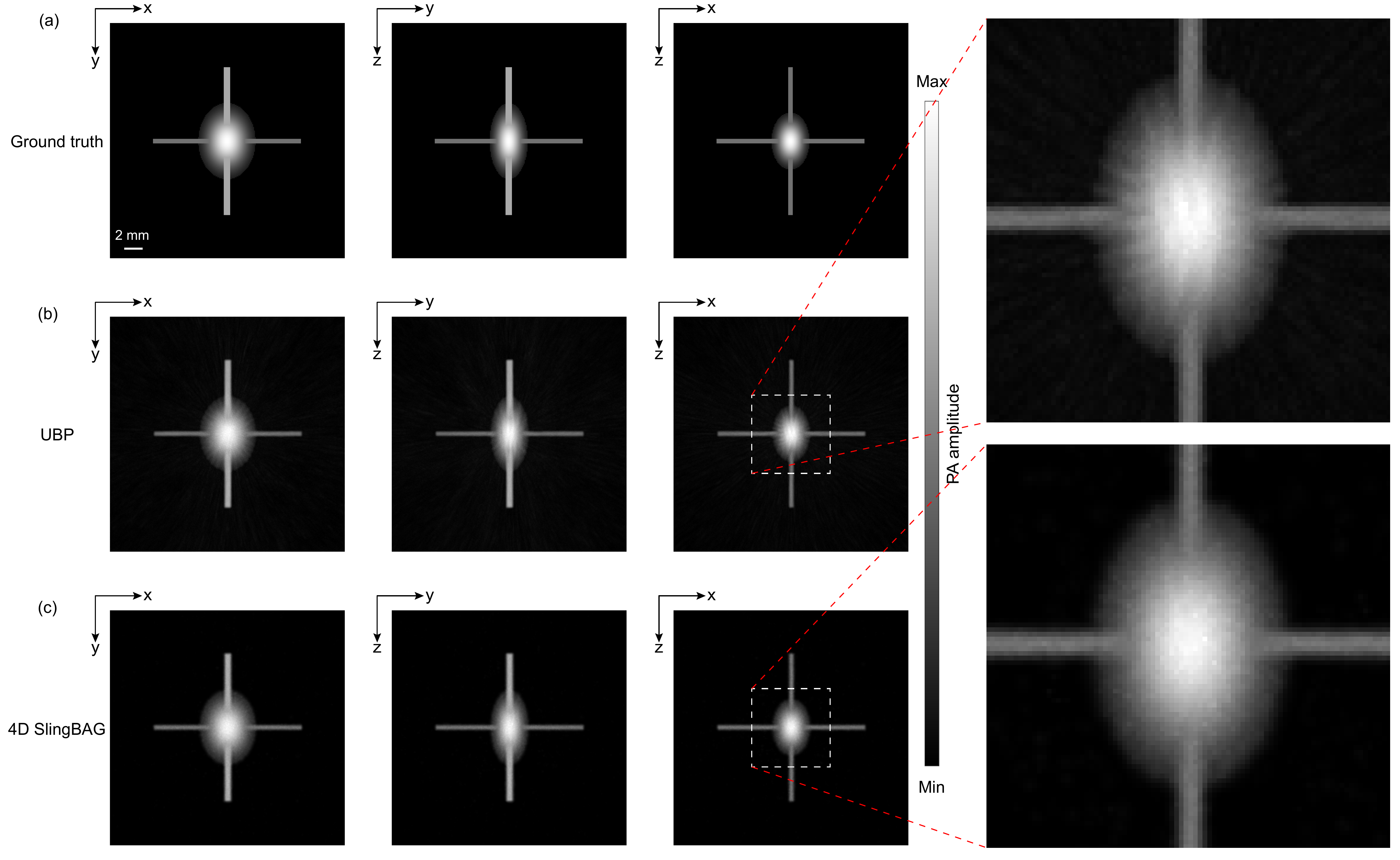}
    \caption{The reconstruction results of the dynamic 3D photoacoustic scene in the 6th frame. (a) XY Plane-MAP, YZ Plane-MAP, XZ Plane-MAP of the ground truth acoustic source. (b) XY Plane-MAP, YZ Plane-MAP, XZ Plane-MAP of the UBP reconstruction results using 1,024 sensor signals. (c) XY Plane-MAP, YZ Plane-MAP, XZ Plane-MAP of the 4D SlingBAG reconstruction results using 1,024 sensor signals. (Scale: 2.0 mm.)}
    \label{fig:fig6}
\end{figure}

\begin{figure}[H]
    \centering
    \includegraphics[width=0.9\linewidth]{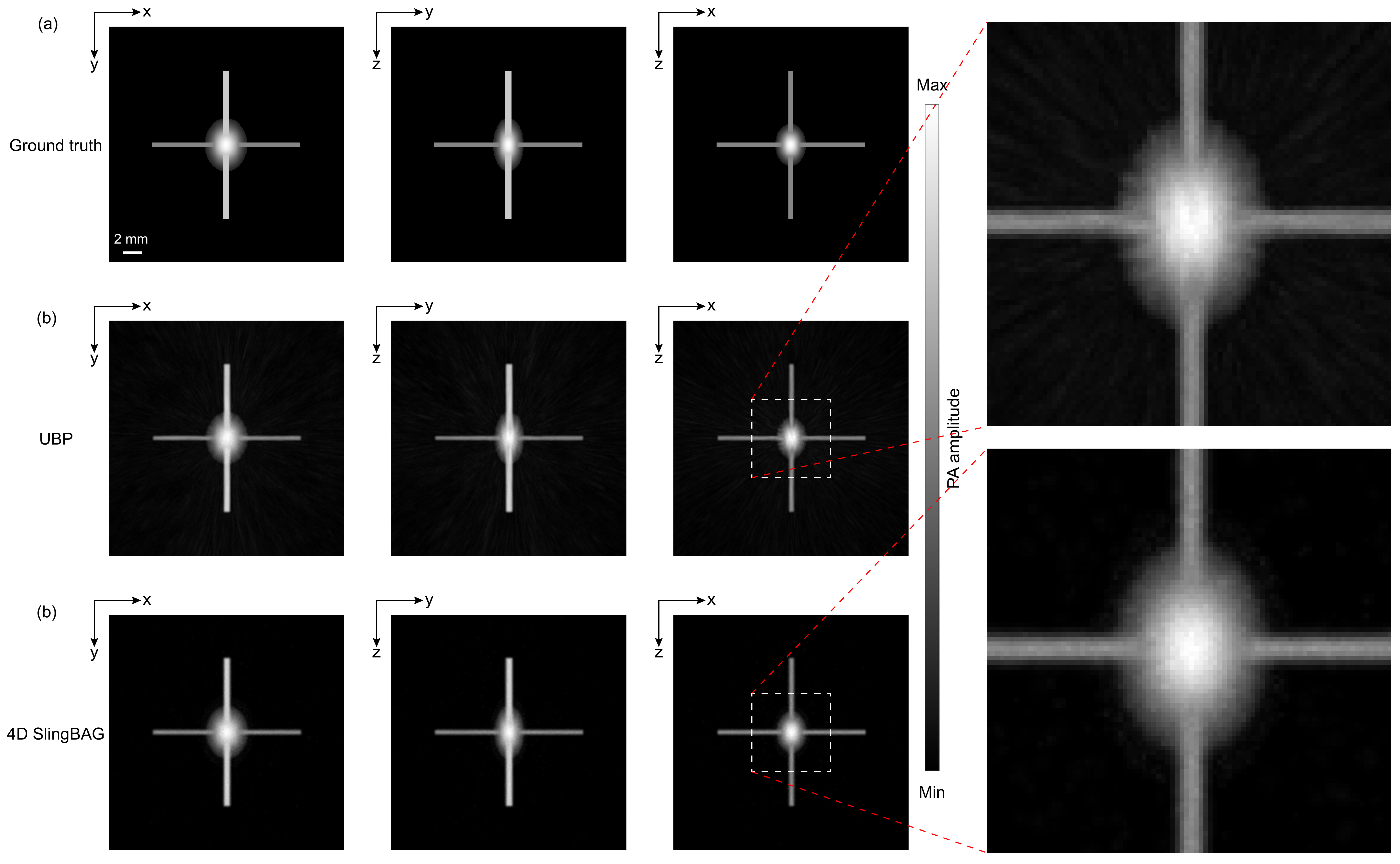}
    \caption{The reconstruction results of the dynamic 3D photoacoustic scene in the 17th frame. (a) XY Plane-MAP, YZ Plane-MAP, XZ Plane-MAP of the ground truth acoustic source. (b) XY Plane-MAP, YZ Plane-MAP, XZ Plane-MAP of the UBP reconstruction results using 1,024 sensor signals. (c) XY Plane-MAP, YZ Plane-MAP, XZ Plane-MAP of the 4D SlingBAG reconstruction results using 1,024 sensor signals. (Scale: 2.0 mm.)}
    \label{fig:fig7}
\end{figure}

We used both UBP algorithm and 4D SlingBAG algorithm to conduct the reconstruction. The results of dynamic 3D PAI reconstruction using the 4D SlingBAG algorithm are shown in the Fig.~\ref{fig:fig4}. We presented the reconstruction results for each frame using maximum amplitude projections (MAP) from a top-down perspective. Additionally, we provided a more detailed comparison of the reconstruction results for the initial frame (Fig.~\ref{fig:fig5}), an intermediate frame (Fig.~\ref{fig:fig6}), and the final frame (Fig.~\ref{fig:fig7}).

It can be observed that the 4D SlingBAG algorithm effectively suppressed artifacts present in the reconstruction results of UBP algorithm under the sparse detector configuration. Furthermore, it significantly reduced distortion in the Gaussian ellipsoid region of the "simulated heart" demonstrating its ability to accurately reconstruct the dynamic PA scene.

\subsection{Simulation experiment 2: complex vascular network reconstruction}

We further validated the 4D SlingBAG algorithm using a more complex dynamic 3D vascular network. The dynamic 3D vascular network was generated based on the reconstruction results of rat liver data~\cite{choi2023deep, choi2023recent, kim20243d, kim2022deep} provided by Professor Chulhong Kim and Dr. Seongwook Choi. Through post-processing steps, we extracted part of the vascular network and simulated a dynamic process involving vascular expansion and enhanced acoustic radiation caused by a strong pulsed light source scanning.

\begin{figure}[ht]
    \centering
    \includegraphics[width=0.9\linewidth]{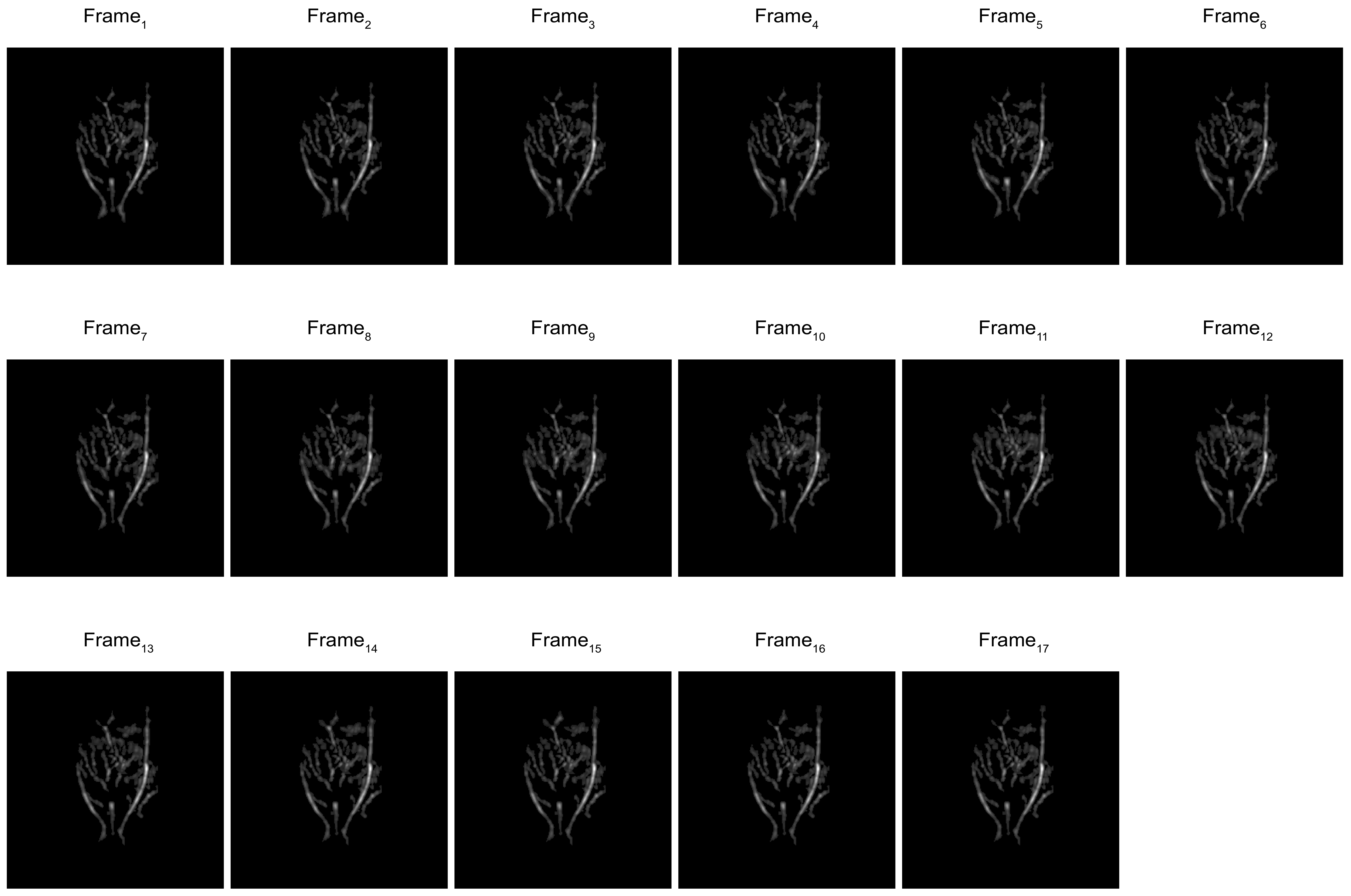}
    \caption{The Top-View-MAP of the dynamic vascular network.}
    \label{fig:fig8}
\end{figure}

The ground truth of the dynamic 3D vascular network is shown in Fig.~\ref{fig:fig8}. The dynamic process consisted of 17 temporal frames, with the vascular network distributed within a cubic region of $x, y, z \in \left(-25.6\text{ mm}, 25.6\text{ mm}\right)$ inside the spherical detector array. The detector array was a spherical array with a radius of 60 mm and comprises 512 detector elements arranged on the sphere following a Fibonacci sequence.

\begin{figure}[ht]
    \centering
    \includegraphics[width=0.9\linewidth]{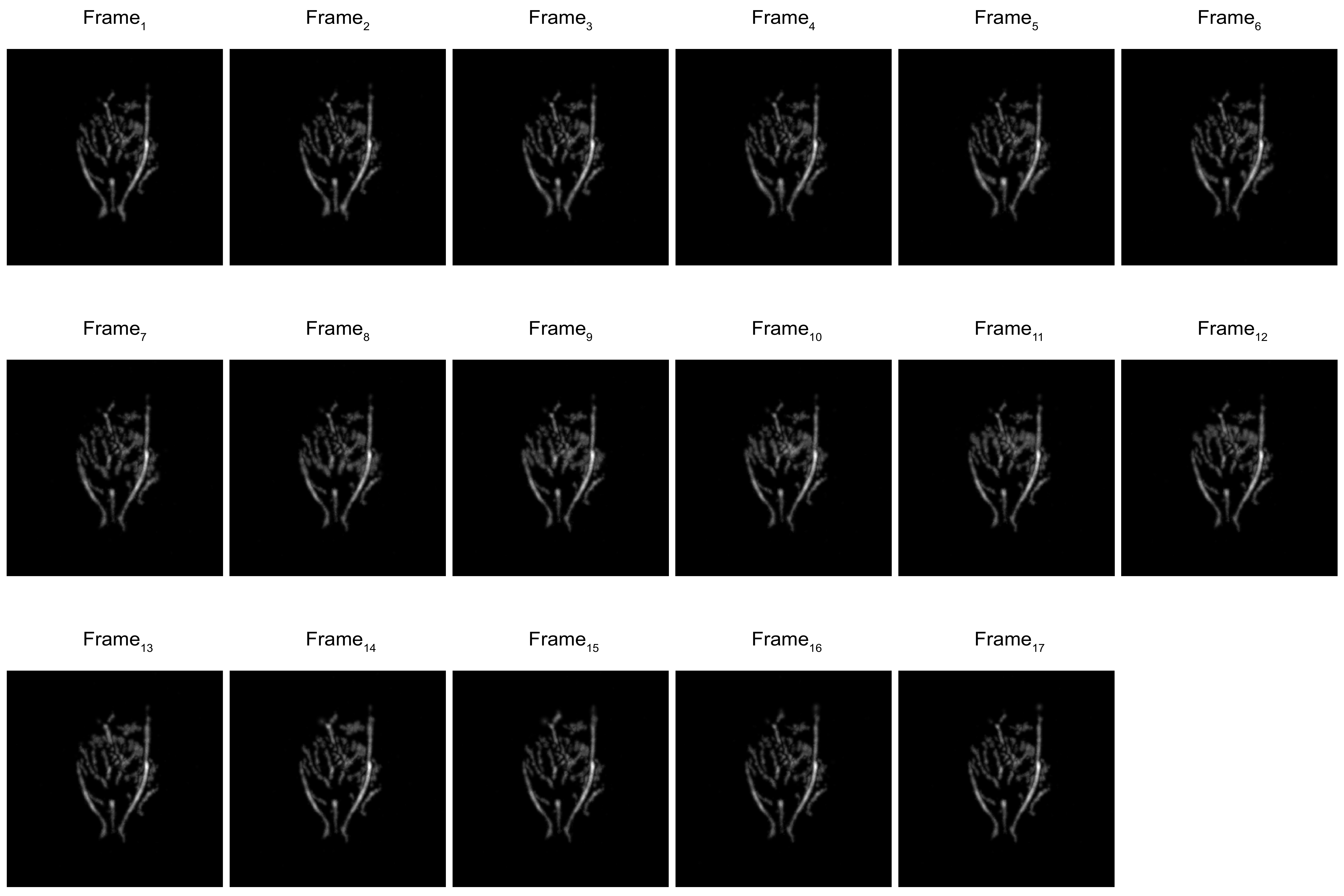}
    \caption{The Top-View-MAP of the reconstruction results of the vascular network using 4D SlingBAG.}
    \label{fig:fig9}
\end{figure}

\begin{figure}[H]
    \centering
    \includegraphics[width=0.9\linewidth]{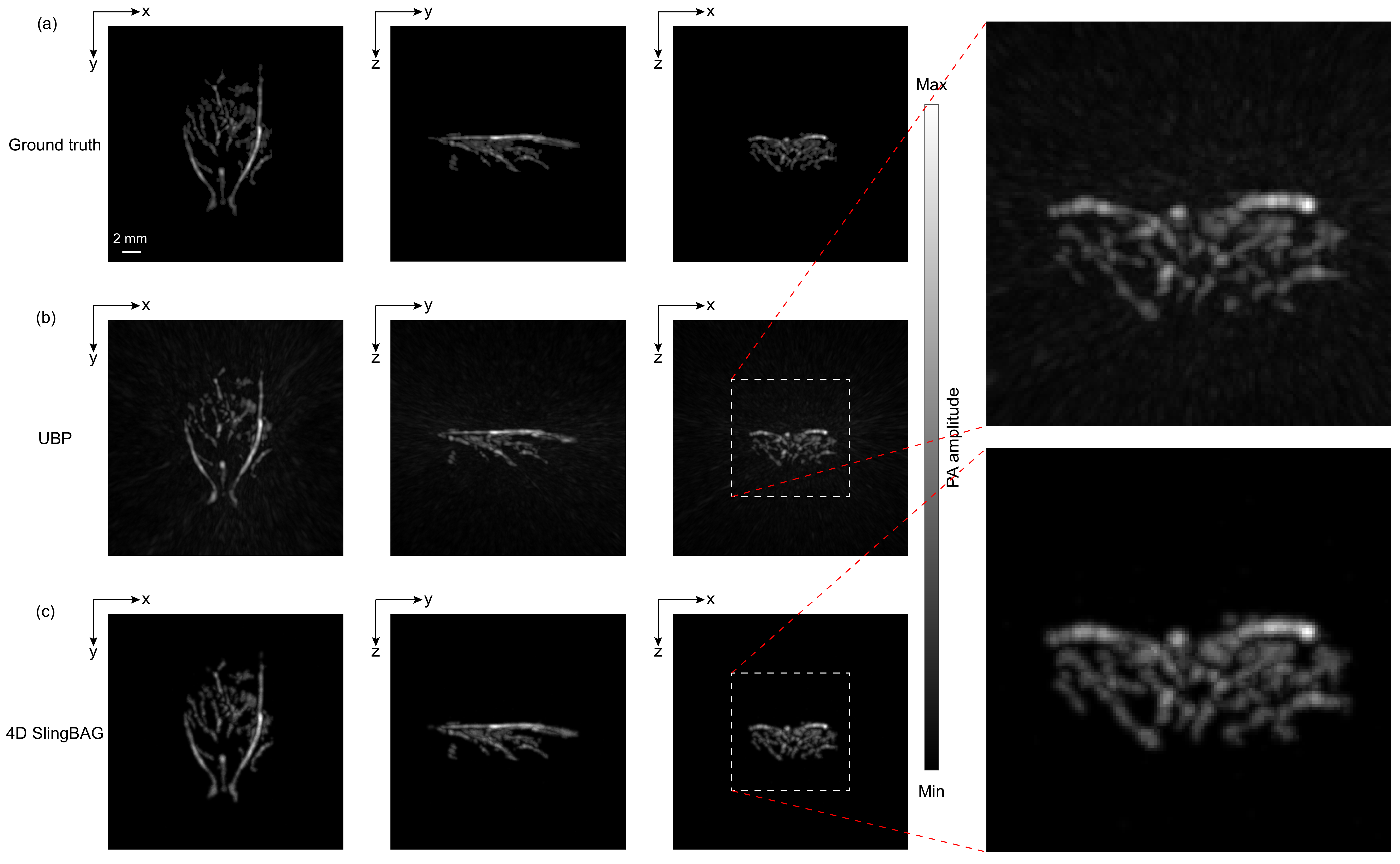}
    \caption{The reconstruction results of the dynamic vascular network in the 1st frame. (a) XY Plane-MAP, YZ Plane-MAP, XZ Plane-MAP of the ground truth acoustic source. (b) XY Plane-MAP, YZ Plane-MAP, XZ Plane-MAP of the UBP reconstruction results using 512 sensor signals. (c) XY Plane-MAP, YZ Plane-MAP, XZ Plane-MAP of the 4D SlingBAG reconstruction results using 512 sensor signals. (Scale: 2.0 mm.)}
    \label{fig:fig10}
\end{figure}

\begin{figure}[H]
    \centering
    \includegraphics[width=0.9\linewidth]{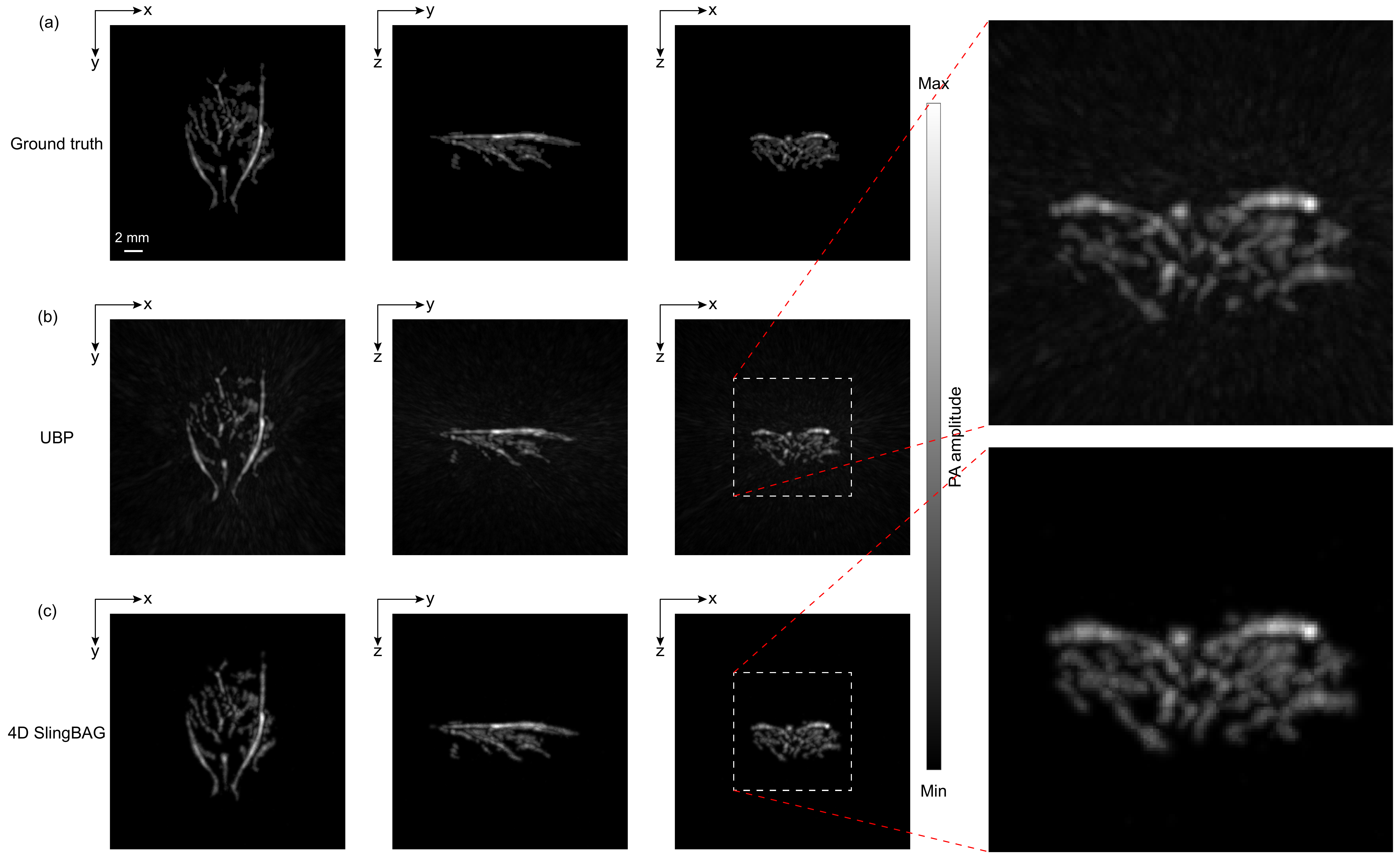}
    \caption{The reconstruction results of the dynamic vascular network in the 6th frame. (a) XY Plane-MAP, YZ Plane-MAP, XZ Plane-MAP of the ground truth acoustic source. (b) XY Plane-MAP, YZ Plane-MAP, XZ Plane-MAP of the UBP reconstruction results using 512 sensor signals. (c) XY Plane-MAP, YZ Plane-MAP, XZ Plane-MAP of the 4D SlingBAG reconstruction results using 512 sensor signals. (Scale: 2.0 mm.)}
    \label{fig:fig11}
\end{figure}

\begin{figure}[H]
    \centering
    \includegraphics[width=0.9\linewidth]{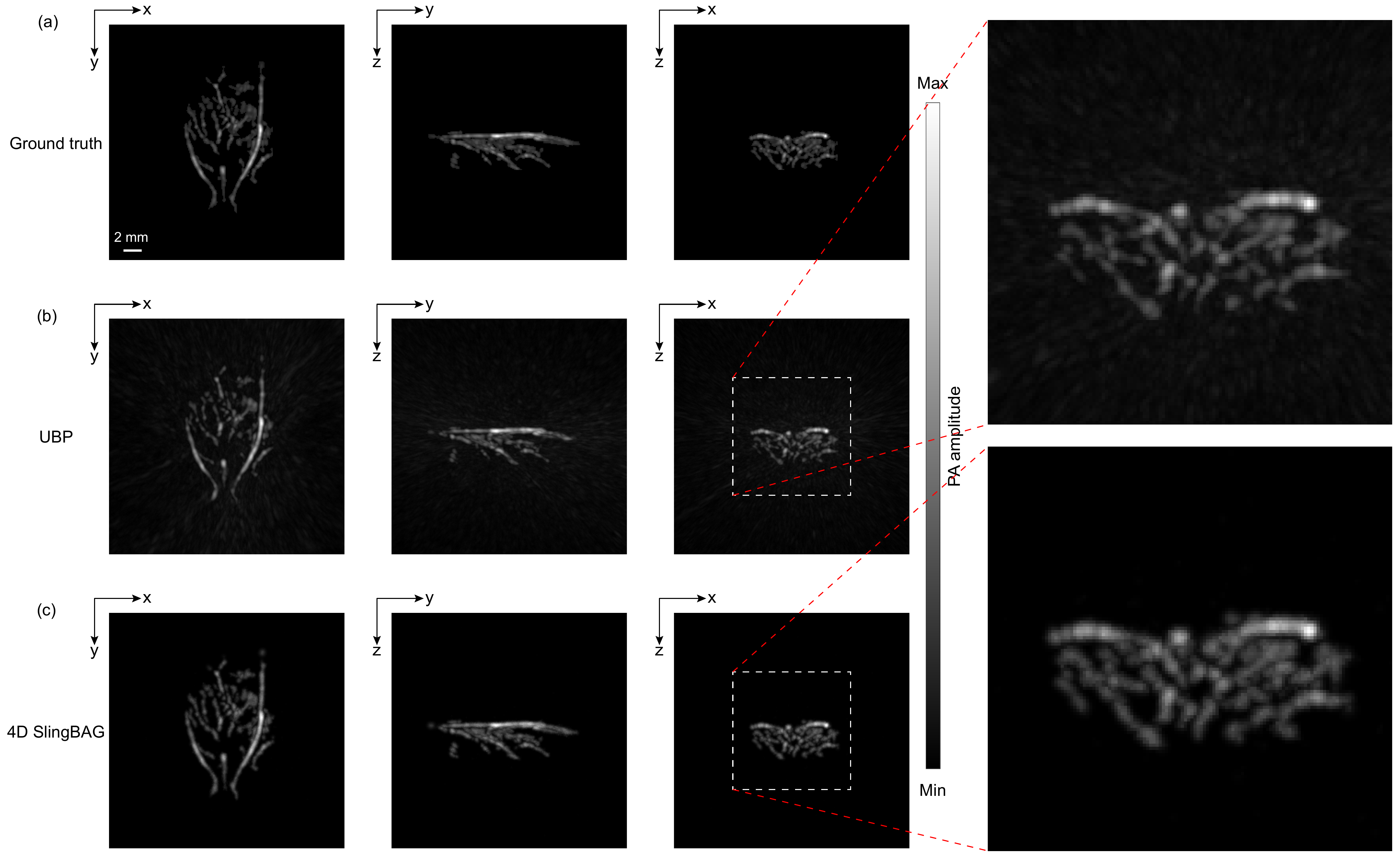}
    \caption{The reconstruction results of the dynamic vascular network in the 17th frame. (a) XY Plane-MAP, YZ Plane-MAP, XZ Plane-MAP of the ground truth acoustic source. (b) XY Plane-MAP, YZ Plane-MAP, XZ Plane-MAP of the UBP reconstruction results using 512 sensor signals. (c) XY Plane-MAP, YZ Plane-MAP, XZ Plane-MAP of the 4D SlingBAG reconstruction results using 512 sensor signals. (Scale: 2.0 mm.)}
    \label{fig:fig12}
\end{figure}

The frame-by-frame reconstruction results of the dynamic 3D vascular network using 4D SlingBAG are shown in the Fig.~\ref{fig:fig9}. We provided a more detailed comparison of the reconstruction results for the initial frame (Fig.~\ref{fig:fig10}), an intermediate frame (Fig.~\ref{fig:fig11}), and the final frame (Fig.~\ref{fig:fig12}). It is evident that the reconstruction results from 4D SlingBAG exhibit significantly less noise compared to UBP and achieve better continuity.

We also computed the Structural Similarity Index (SSIM) between the frame-by-frame reconstruction results and the ground truth for the maximum amplitude projection (MAP) images in the XY-plane, YZ-plane, and XZ-plane (see Tab.~\ref{tab:ssim_results}). The results demonstrated that, under the sparse 512-element spherical array configuration, 4D SlingBAG achieved high-quality 4D reconstruction, with the SSIM values for the MAP images in all three directions exceeding 0.93. In contrast, under the same conditions, the UBP reconstruction results delivered SSIM values below 0.1. These findings confirm the significant advantage of 4D SlingBAG for dynamic 3D PAI reconstruction.

\begin{table}[ht]
\centering
\caption{SSIM of reconstruction results of complex vascular network under
different MAP views}
\label{tab:ssim_results}
\[
\begin{array}{c|ccc|ccc}
\hline
\textnormal{Algorithms} & \multicolumn{3}{c|}{\textbf{UBP}} & \multicolumn{3}{c}{\textbf{4D SlingBAG}} \\ \hline
\textnormal{View of MAP} & \textbf{XY-Plane} & \textbf{YZ-Plane} & \textbf{XZ-Plane} & \textbf{XY-Plane} & \textbf{YZ-Plane} & \textbf{XZ-Plane} \\ \hline
\textnormal{frame 1} & {\scriptstyle 0.0823} & {\scriptstyle 0.0528} & {\scriptstyle 0.0501} & {\scriptstyle 0.9329} & {\scriptstyle 0.9615} & {\scriptstyle 0.9738} \\
\textnormal{frame 2} & {\scriptstyle 0.0827} & {\scriptstyle 0.0529} & {\scriptstyle 0.0503} & {\scriptstyle 0.9308} & {\scriptstyle 0.9606} & {\scriptstyle 0.9724} \\
\textnormal{frame 3} & {\scriptstyle 0.0829} & {\scriptstyle 0.0529} & {\scriptstyle 0.0508} & {\scriptstyle 0.9321} & {\scriptstyle 0.9604} & {\scriptstyle 0.9730} \\
\textnormal{frame 4} & {\scriptstyle 0.0834} & {\scriptstyle 0.0540} & {\scriptstyle 0.0509} & {\scriptstyle 0.9322} & {\scriptstyle 0.9602} & {\scriptstyle 0.9723} \\
\textnormal{frame 5} & {\scriptstyle 0.0834} & {\scriptstyle 0.0538} & {\scriptstyle 0.0512} & {\scriptstyle 0.9310} & {\scriptstyle 0.9606} & {\scriptstyle 0.9724} \\
\textnormal{frame 6} & {\scriptstyle 0.0837} & {\scriptstyle 0.0540} & {\scriptstyle 0.0517} & {\scriptstyle 0.9304} & {\scriptstyle 0.9610} & {\scriptstyle 0.9726} \\
\textnormal{frame 7} & {\scriptstyle 0.0837} & {\scriptstyle 0.0539} & {\scriptstyle 0.0512} & {\scriptstyle 0.9296} & {\scriptstyle 0.9602} & {\scriptstyle 0.9710} \\
\textnormal{frame 8} & {\scriptstyle 0.0840} & {\scriptstyle 0.0532} & {\scriptstyle 0.0517} & {\scriptstyle 0.9319} & {\scriptstyle 0.9612} & {\scriptstyle 0.9728} \\
\textnormal{frame 9} & {\scriptstyle 0.0838} & {\scriptstyle 0.0528} & {\scriptstyle 0.0517} & {\scriptstyle 0.9316} & {\scriptstyle 0.9608} & {\scriptstyle 0.9724} \\
\textnormal{frame 10} & {\scriptstyle 0.0839} & {\scriptstyle 0.0529} & {\scriptstyle 0.0509} & {\scriptstyle 0.9322} & {\scriptstyle 0.9596} & {\scriptstyle 0.9721} \\
\textnormal{frame 11} & {\scriptstyle 0.0838} & {\scriptstyle 0.0532} & {\scriptstyle 0.0509} & {\scriptstyle 0.9320} & {\scriptstyle 0.9606} & {\scriptstyle 0.9727} \\
\textnormal{frame 12} & {\scriptstyle 0.0831} & {\scriptstyle 0.0519} & {\scriptstyle 0.0493} & {\scriptstyle 0.9321} & {\scriptstyle 0.9605} & {\scriptstyle 0.9734} \\
\textnormal{frame 13} & {\scriptstyle 0.0834} & {\scriptstyle 0.0530} & {\scriptstyle 0.0502} & {\scriptstyle 0.9323} & {\scriptstyle 0.9613} & {\scriptstyle 0.9727} \\
\textnormal{frame 14} & {\scriptstyle 0.0833} & {\scriptstyle 0.0534} & {\scriptstyle 0.0506} & {\scriptstyle 0.9311} & {\scriptstyle 0.9602} & {\scriptstyle 0.9724} \\
\textnormal{frame 15} & {\scriptstyle 0.0829} & {\scriptstyle 0.0529} & {\scriptstyle 0.0506} & {\scriptstyle 0.9312} & {\scriptstyle 0.9609} & {\scriptstyle 0.9737} \\
\textnormal{frame 16} & {\scriptstyle 0.0817} & {\scriptstyle 0.0525} & {\scriptstyle 0.0501} & {\scriptstyle 0.9310} & {\scriptstyle 0.9599} & {\scriptstyle 0.9723} \\
\textnormal{frame 17} & {\scriptstyle 0.0809} & {\scriptstyle 0.0523} & {\scriptstyle 0.0501} & {\scriptstyle 0.9321} & {\scriptstyle 0.9609} & {\scriptstyle 0.9733} \\
\hline
\end{array}
\]
\end{table}

The reconstruction of a single reference frame using SlingBAG took approximately 2.85 hours, and the subsequent dynamic 3D reconstruction using 4D SlingBAG required an additional 3.13 hours. The total runtime was less than 6 hours, which represents a more than 7-fold reduction in time compared to reconstructing dynamic frames independently with SlingBAG. This demonstrates the exceptional efficiency of 4D SlingBAG for large-scale dynamic 3D PAI reconstruction. In addition, the memory consumption for dynamic 3D reconstruction using 4D SlingBAG is almost the same as that for single-frame 3D reconstruction with SlingBAG.

4D SlingBAG applies 65 learnable Gaussian basis functions to construct the spatial-temporal coupled deformation function. The initial learning rates for the 3D coordinates, peak initial pressure, standard deviation, and deformation coefficients were set to $5 \times 10^{-7}$, $5 \times 10^{-4}$, $5 \times 10^{-7}$, and $5 \times 10^{-6}$, respectively, with a step decay of $step\_size = 160$ and $drop\_rate = 0.1$. All experiments were conducted on a single NVIDIA RTX 3090 Ti GPU.

\section{Discussion}

High-quality large-scale dynamic 3D photoacoustic imaging (PAI) reconstruction has always been a critical topic and holds great significance for the clinical application. The 4D SlingBAG algorithm builds upon the iterative point cloud framework of SlingBAG and incorporates spatial-temporal coupled deformation functions to enable efficient dynamic 3D PAI reconstruction. It achieves high-quality reconstruction of large-scale dynamic PA scenes with extremely fast reconstruction speed and low memory consumption. However, 4D SlingBAG does not currently account for conditions with acoustic velocity inhomogeneities and tissue acoustic attenuation. In future work, we aim to address this limitation by employing learnable spherical harmonic (SH) functions to model the anisotropic average acoustic speed required for each 4D Gaussian sphere's computation due to speed inhomogeneities.

The 4D SlingBAG algorithm is highly versatile and can be easily adapted to various detector array configurations, making it a powerful tool for large-scale dynamic 3D PAI reconstruction. It is expected to play a significant role in fast hemodynamic change scenarios, such as drug delivery and neuronal metabolism studies.

\section*{Acknowledgments}

This research was supported by the following grants: the National Key R\&D Program of China (No. 2023YFC2411700, No. 2017YFE0104200); the Beijing Natural Science Foundation (No. 7232177); the National Natural Science Foundation of China (62441204, 62472213); the Basic Science Research Program through the National Research Foundation of Korea (NRF) funded by the Ministry of Education (2020R1A6A1A03047902).

\bibliographystyle{ieeetr}
\bibliography{references}

\end{document}